\documentclass{article}

\PassOptionsToPackage{numbers}{natbib}

\usepackage[final]{nips_2017}

\usepackage[utf8]{inputenc} 
\usepackage[T1]{fontenc}    
\usepackage{hyperref}       
\usepackage{url}            
\usepackage{booktabs}       
\usepackage{amsfonts}       
\usepackage{nicefrac}       
\usepackage{microtype}      
\usepackage{graphicx}
\usepackage{subcaption}
\usepackage{amsmath}
\usepackage{amssymb}
\usepackage{textcomp}
\usepackage{gensymb}

\usepackage{xcolor}
\usepackage{soul}

\title{One-Shot Reinforcement Learning for Robot Navigation
with Interactive Replay}

\author{
  Jake Bruce\\
  QUT, Brisbane\\
  \texttt{jacob.bruce@hdr.qut.edu.au}\\
  \And
  Niko Sünderhauf\\
  QUT, Brisbane\\
  \texttt{niko.suenderhauf@qut.edu.au}\\
  \And
  Piotr Mirowski\\
  DeepMind, London\\
  \texttt{piotrmirowski@google.com}\\
  \And
  Raia Hadsell\\
  DeepMind, London\\
  \texttt{raia@google.com}\\
  \And
  Michael Milford\\
  QUT, Brisbane\\
  \texttt{michael.milford@qut.edu.au}\\
}

\begin{document}

\maketitle

\begin{abstract}
Recently, model-free reinforcement learning algorithms have been shown to solve challenging
problems by learning from extensive interaction with the environment. A significant issue with
transferring this success to the robotics domain is that interaction with the real world is costly,
but training on limited experience is prone to overfitting. We present a method for learning to
navigate, to a fixed goal and in a known environment, on a mobile robot. The robot leverages an interactive world model built from a single traversal
of the environment, a pre-trained visual feature encoder, and stochastic environmental augmentation,
to demonstrate successful zero-shot transfer under real-world environmental variations without
fine-tuning.
\end{abstract}

\section{Introduction}

Learning from experience is an important capability with potential implications in all areas of
robotics. However, interaction with the world can be an expensive undertaking due to constraints such
as power usage and human supervision. As a result, a priority for any robotic learning
system is minimization of the amount of environmental interaction required to learn a task. Model-free
reinforcement learning systems have been shown to solve complex Markov decision processes (MDPs) in a
variety of challenging domains, but usually at the cost of a large amount of agent experience
(not to be confused with the number of training steps required by the optimization method).
In this work we learn to reliably navigate towards a fixed goal in a known, real world environment, using \emph{interactive replay}
of a single traversal of the environment. Additional contributions of this paper include using
pre-trained visual features, and augmenting the training set with stochastic observations,
to demonstrate zero-shot transfer to real-world variations in the environment unseen during training.

To collect enough raw experience to train a model-free system on a real-world task is
prohibitively expensive or undesirable in many robotics situations;
and a common limitation of mobile robots is limited battery capacity with long recharge
times, in which the physical platform is constrained but computational resources are unoccupied.
To exploit these limitations and opportunities inherent in mobile robotics,
we propose to learn offline using interactive replay from an
internal world model~\cite{vaughan2006use} that we generate from a single traversal of the environment.
In order to combat the tendency to overfit on such a small training environment we make
use of a pre-trained and fixed visual feature encoder, and we apply stochastic
augmentations to the environment; these modifications enable zero-shot transfer
to variations in the environment unseen by the agent during training.
Our system differs from previous work on learning to navigate from real-world
imagery~\cite{zhu2017target} in that our environments are an order of magnitude larger,
and we transfer to a validation environment gathered under natural everyday variation
with no fine-tuning required.

\begin{figure}[t]
    \centering
    \begin{subfigure}{0.3435\linewidth}
        \centering
        \includegraphics[width=\linewidth]{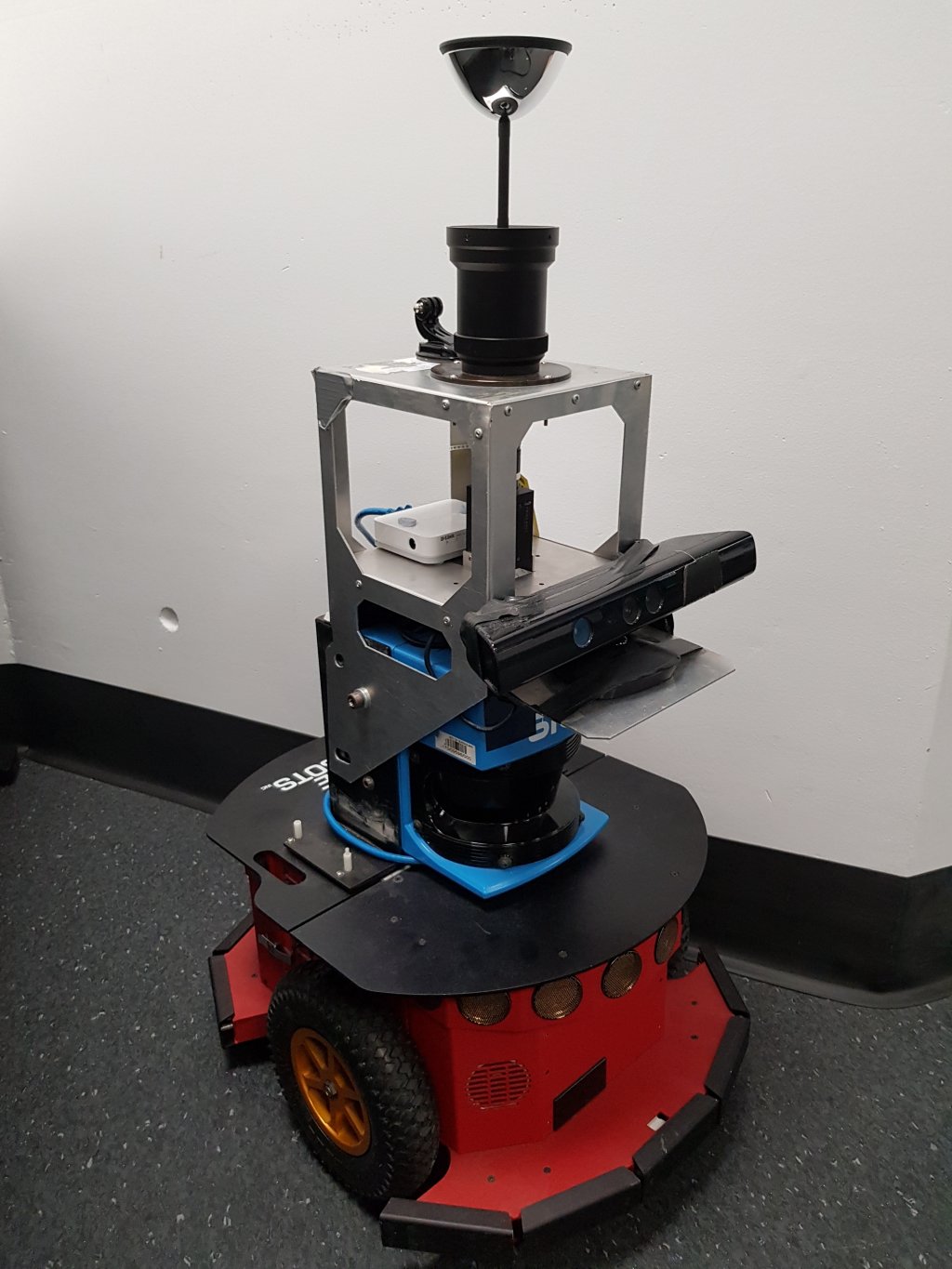}
        \caption{}
        \label{fig:pioneer}
    \end{subfigure}\hfill%
    \begin{subfigure}{0.62\linewidth}
        \centering
        \includegraphics[width=\linewidth]{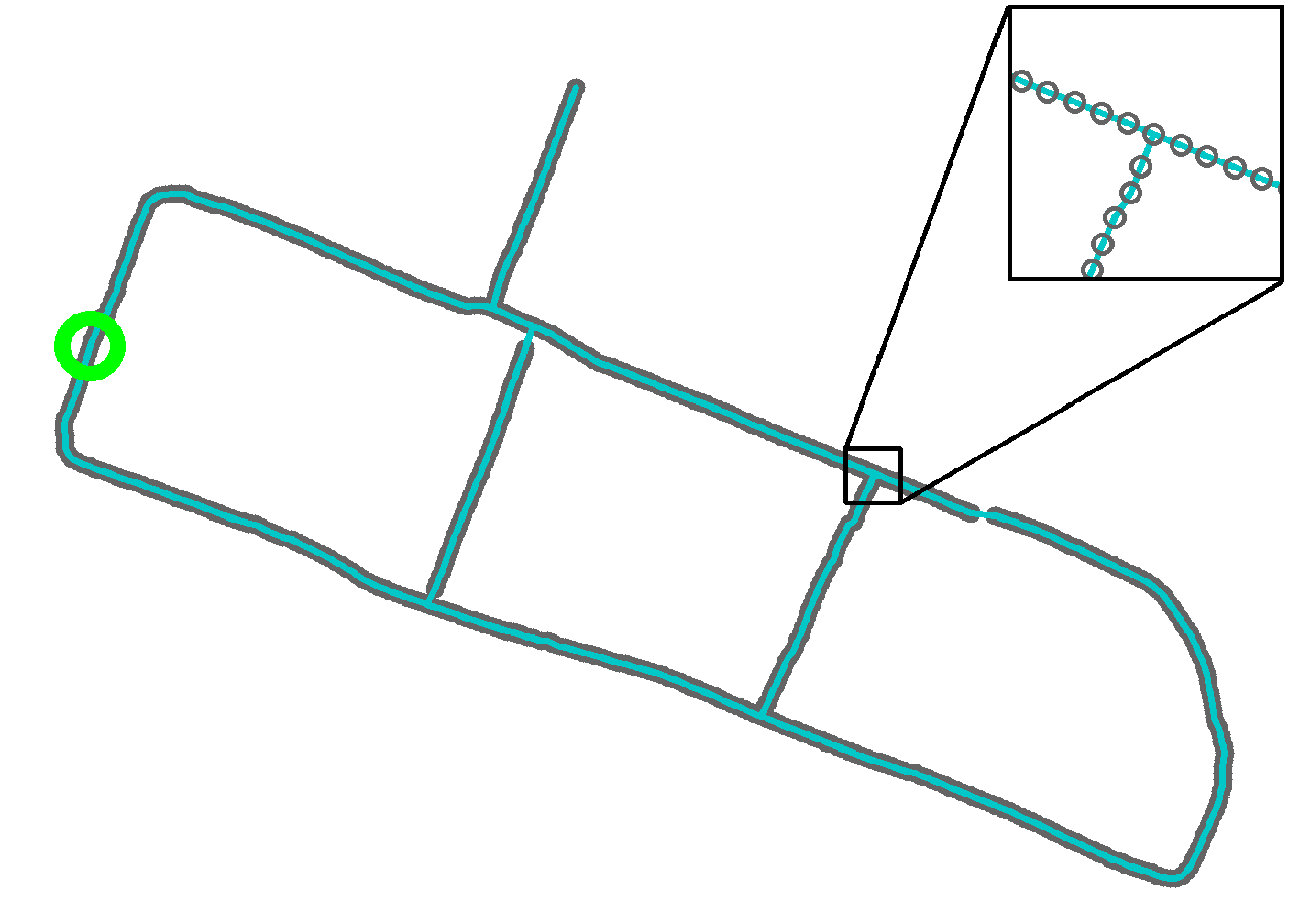}
        \caption{}
        \label{fig:environment_map}
    \end{subfigure}

    \caption{a) Pioneer 3DX mobile robot used in our experiments. b) Graph of the environment
    built from a single traversal; green circle indicates goal location, and inset shows local connection structure.}
    \label{fig:environment}
\end{figure}

\section{Related Work}

In this section, we briefly review reinforcement learning in general,
challenging problems in which it has achieved recent success, and the specific
problem of navigation.

\subsection{Reinforcement Learning}

We formulate the navigation problem as an MDP described by the tuple
$(\mathcal{S},\mathcal{A},\mathcal{T},\mathcal{O},\mathcal{F})$:
$\mathcal{S}$~is the set of agent-environment states,
$\mathcal{A}$~is the set of possible actions the agent can perform,
$\mathcal{T}(s,a) \rightarrow p(s^\prime|s,a)$~maps a state-action pair to the next state,
$\mathcal{O}(s) \rightarrow p(x|s)$~generates agent-visible observations given the true state,
and
the reward function $\mathcal{F}(s) \rightarrow r$~describes whether the agent has achieved the goal.

Note that the formulation of our MDP allows the observation function
$\mathcal{O}(s)$ to be non-deterministic, and demonstrating the transfer performance implications of stochastic observations is a key contribution of this work.
The transition function $\mathcal{T}(s,a)$ can also be stochastic, but we consider
deterministic transitions for simplicity.

Reinforcement learning (RL) involves finding a policy $\pi(x) \rightarrow a$ to map observations to actions that
maximize the expected sum of future rewards, often referred to as the \emph{return}, denoted by $R$.
The return can take many forms, but a common choice is the sum of $\gamma$-discounted future rewards
$R=\sum_t \gamma^t r_t$. Broad flavors of RL include model-based control in which a predictive model is used to plan actions or learn a policy, episodic control in which highly rewarding trajectories can be replayed from known states, and model-free control in which a policy is learned directly from observations. We consider the model-free setting in this work.

In model-free RL, the policy is implemented by a function approximator, commonly a
deep neural network. The system is usually trained either to map observations and
actions to the expected future reward (as in Q-learning) or to directly approximate
the probability distribution over actions that maximizes
expected reward (as in policy search). When using a deep neural network as a function
approximator, a policy is
learned from experience with the environment, usually by gradient descent on
trajectories of $(x,a,R)$-tuples. In value-based methods
such as Q-learning, the policy consists of using a value estimate (the $Q$ function)
to choose actions with the highest return, whereas in
policy search methods, the policy is parameterized directly and a value function is used to
update the parameters of the policy in the direction of positive outcomes.
We primarily consider a value-based method known as \emph{bootstrapped Q-learning}~\cite{osband2016deep}.

\subsection{Recent Successes}

Many challenging problems have recently been addressed with model-free RL, with a
particular abundance of research on the topic of playing video games directly
from pixels. Video games make effective testbeds for evaluating
learning algorithms due to full observability and manipulability, the ability to run
at high speeds to gather large amounts of agent experience, and the potential
for structural and perceptual similarities to real-world problems.
Recent examples include learning to play a diverse suite of dozens of low-resolution
Atari games with deep Q-networks~\cite{mnih2015human}, challenging board games such as
Go~\cite{silver2016mastering}, and first-person shooters such as Doom~\cite{kempka2016vizdoom,dosovitskiy2016learning}.

Real world tasks have also been a target for RL algorithms, especially for
difficult perception and control problems in robotics~\cite{kober2013reinforcement}.
Recent contributions include continuous control for simulated robots~\cite{duan2016benchmarking},
motor control for manipulation directly from pixels~\cite{levine2016end}, and simulation-to-real
transfer for manipulation~\cite{zhang2015towards,rusu2016sim}.

The successful application of deep Q-networks~\cite{mnih2015human} involved the
use of a passive replay buffer of past experience, which enabled more efficient use of
interaction samples and stabilized training. We build on the concept in this work by introducing
\emph{interactive replay}, where a single trajectory through a real environment
comprises a virtual environment through which an agent can learn to navigate.

\subsection{Reinforcement Learning for Navigation}

Recent work has also addressed the task of navigating 3D environments
in simulation, e.g. maze navigation from perspective view
imagery~\cite{mnih2016asynchronous,mirowski2016learning}. In robotics, agents have been trained
to navigate using laser sensors in simulation~\cite{tai2017virtual} and simulated depth
imagery~\cite{zhang2016deep}; structure-based sensors tend to exhibit less difference
between simulation and reality (a narrower ``reality gap'') than does visual sensing.
However, visual navigation is also an established direction, including work on target-driven
image-based navigation in small grid worlds with simulation and fine-tuning in the real
world~\cite{zhu2017target}.

In this work, we build on the success of navigation learning in virtual
environments~\cite{mirowski2016learning} and preliminary work in the real
world~\cite{zhu2017target} to develop techniques for learning to navigate large office environments
on a mobile robot, with goal-generated reward as the only supervised training signal.

\section{Approach}

In this section we describe our approach consisting of the construction of virtual
training and validation environments from single traversals of the real world; our use
of a pre-trained and fixed visual encoder; stochastic observations to augment the
training environment; and the bootstrapped Q-learning algorithm for RL.

\subsection{Interactive Replay}

The success of deep Q-networks~\cite{mnih2015human} can be attributed partially to the
use of a passive replay memory of past experience, from which training batches are sampled
in order to make more efficient use of interaction samples and to stabilize training. We
build on the concept of experience replay~\cite{mcclelland1995there,lin1993reinforcement}, and more specifically that of a replay buffer, by introducing \emph{interactive replay},
in which a rough world model is memorized from a single traversal of the environment.
This allows an agent to interact with the model to generate a large number of diverse trajectories
for learning to navigate, while minimizing the amount of real-world experience required
by the robot.

In this section we describe how the training environment is gathered in one pass, and
minimal human annotation is used to construct a pose graph for virtual interactive replay;
a validation environment is then constructed from a second pass under different environmental
conditions and approximately aligned with the training environment.

\paragraph{Training environment}


Recording the training environment consists of two phases. First, the robot executes a complete
traversal of the environment while recording sensor data. Second, the recorded data is aligned
into a topological map of the environment. In our experiments, alignment is conducted by minimal human
interaction in which the user is shown
the approximate poses of the agent in two-dimensional space (see Fig.~\ref{fig:environment_map}),
connects loops that were closed during the traversal, and removes
duplicated regions of the space. In the general case, both phases can be accomplished autonomously
using simultaneous localization and mapping (SLAM) techniques~\cite{dissanayake2001solution}
without human intervention.

The space is then discretized into a pose graph~\cite{thrun2006graph} consisting of sensory snapshots taken every
$\Delta_\text{pos}$ meters, which are connected by edges representing temporal adjacency and loop closures.
In the office environment considered in this work, the agent can rotate in $\Delta_\text{rot}$ increments
in either direction, and move forward by $\Delta_\text{pos}$ (if there is no node in that direction,
the action has no effect). Nodes in the pose graph may have
edges along the forward, backward, left, and right directions, but the pose graph concept
generalizes to environments that are not grid-like. The resulting graph comprises the
training environment in which the agent will learn to navigate.

\paragraph{Validation environment}

In order to evaluate the transfer performance of the agent under environmental variations that
it has not seen during training, we gather a separate environment to use as validation during
training. Collection of the validation environment is simpler than the training environment: the
robot is used to conduct a second traversal of the environment, not necessarily in the same order
as the first. Then the stored training environment is used in conjunction with an approximate
localization method to associate nodes in the pose graph of the training environment with the
closest matching pose in the validation recording; in our work we use laser-based localization
as an approximate localization system and a $360\degree$ camera to align
the orientation of the imagery, but the pose correspondences can be annotated by hand if a
localization system is not available. The resulting aligned graph comprises a validation environment
to estimate transfer performance under realistic environmental variation.

%

\subsection{Rich Visual Encoding}

A significant issue with training on a small set of real-world experience is the tendency
to overfit to this constrained environment. One area where overfitting can manifest
is the learning of visual features, since most current approaches in model-free RL learn
a feature extractor from scratch~\cite{mnih2015human,mnih2016asynchronous,mirowski2016learning},
and a single traversal of a small environment provides limited variety of visual structure.
Using a visual encoder network that has been trained on a large vision dataset can help
compensate for the lack of variety in the training environment~\cite{sunderhauf2015place},
so we use a ResNet-50 network trained on ImageNet~\cite{he2016deep,deng2009imagenet} as in~\cite{zhu2017target}
and extract the pre-final fully-connected layer as a $2048$-dimensional rich visual encoding
of each of the four $90\degree$ views from the camera. These are concatenated into an
$8192$-dimensional vector to form the agent's observation $f(x_t)$.
We also freeze the weights on this encoder to avoid
overfitting to the limited structure present in the training environment; this also allows
us to pre-compute the features in our recordings, yielding a significant computational
advantage during training.

\subsection{Stochastic Observations}

Data augmentation techniques have been established to compensate for limited training data,
in which random transformations are applied to the training examples in order to improve the
diversity of the training samples~\cite{gan2015learning}.
We leverage this idea in our training environment by manipulating the observation
function $\mathcal{O}(s) \rightarrow p(x|s)$ of our MDP to return observations drawn from a
distribution of positions and orientations centered on the true pose of the state $s$.
In this work, we use the nodes in the environment's pose graph as the set of states
$\mathcal{S}$, and generate observations by sampling from a distribution over both position
and orientation.


For position, we sample input observations $x_{t}$ from images recorded
``in between'' the discretized nodes of the pose graph according to a normal
distribution centered on the true state $s_{t}$ with a variance of $5$cm. Orientations
are sampled from a normal distribution centered on the true orientation of $s_{t}$, with a
variance of $5\degree$; the distributions approximate the accuracy of our laser-based
localization system.
In Section~\ref{sec:experiments} we investigate the degree to which a stochastic
observation function reduces overfitting in our navigation problem, as measured by relative
performance on the validation set.

\subsection{Bootstrapped Q-Learning}

We consider as our model-free RL algorithm a double dueling $n$-step Q-learning
framework~\cite{van2016deep,wang2015dueling,mnih2016asynchronous} based on~\cite{osband2016deep}
with $N_Q$ parallel Q-function \emph{heads}. Each head is a deep neural network with shared
visual encoder and recurrent layers, but with its own Q-function output layers.
One head is chosen at  random as the behavior generator each episode, and each head is
trained on a fraction $p_Q$ of the data of all the heads (see~\cite{osband2016deep} for details).
We refer to the architecture as \emph{Bootstrapped Q-learning}.

The bootstrapped Q-network is trained by minimizing the traditional Q-learning loss function: $\mathcal{L} = \sum\limits_{\text{batch}} (\overset{\sim}{Q}(f(x_{t})) - R_{t})^2$, where $R_{t}$ is the
exponentially-discounted sum of future rewards.
As in~\cite{mnih2016asynchronous}, we run $N_w$ parallel workers exploring the environment
simultaneously using a single shared network, which is updated by stochastic gradient descent
on the batches of experience composed of all of the workers' experience. Note that this parallelism
is distinct from the $N_Q$ separate Q-function heads, which comprise an additional level of
parallel structured exploration on top of the parallel workers. In this work, we use $N_w=64$,
$N_Q=10$, and $p_Q=0.5$.

\begin{figure}[t]
    \centering
    \includegraphics[width=0.85\linewidth]{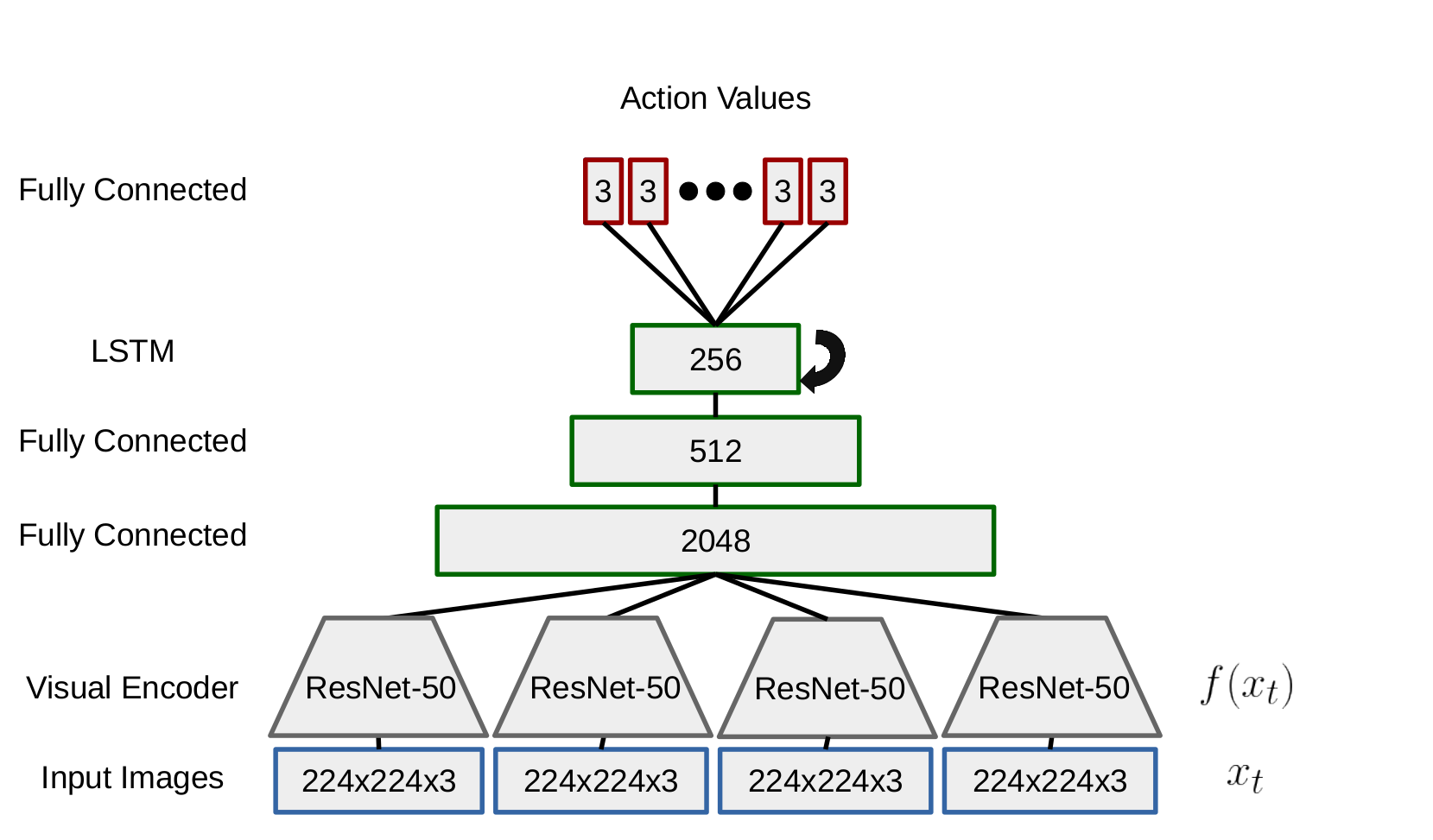}
    \caption{Schematic of our model architecture. Blue outlines indicate inputs; gray outlines indicate
    modules with fixed weights; green outlines indicate layers with trainable weights; and red outlines indicate the $N_Q$ parallel Q-function output layers. The agent sees in four directions at a time, each consisting of a $90\degree$ crop from the robot's $360\degree$ camera.}
    \label{fig:model}
\end{figure}

For an intuitive explanation of the advantage of the bootstrapped approach on our problem,
consider performing a random walk.
In our environment, as in most robot navigation problems, the optimal
prior over actions heavily favors moving forward. By sampling from
a diverse set of random priors in the form of randomly-initialized Q-functions, the agent can
more reliably encounter trajectories that result from near-optimal priors, enabling
faster convergence. As a bonus for the robotics context in particular, the ensemble nature
of the technique enables the interpretation of
the distinct Q-estimates as a probability distribution, enabling reasoning about the
uncertainty in the model. This provides a significant practical advantage over most RL algorithms
(although, see~\cite{bellemare2017distributional} for an exception).
The architecture of the network is shown in Fig.~\ref{fig:model} and corresponds to a standard convolutional recurrent network used in previous work~\cite{mnih2016asynchronous,jaderberg2016reinforcement,mirowski2016learning}. We opted for the use of a recurrent network (specifically a Long Short-Term Memory~\cite{hochreiter1997long}) as policy inputs, instead of a plain feed-forward network, because the recurrent network enables the agent to accumulate visual evidence over time.

\section{Experiments}
\label{sec:experiments}

\begin{figure}[t]
    \centering
    \begin{subfigure}{0.495\linewidth}
        \includegraphics[width=1.0\linewidth]{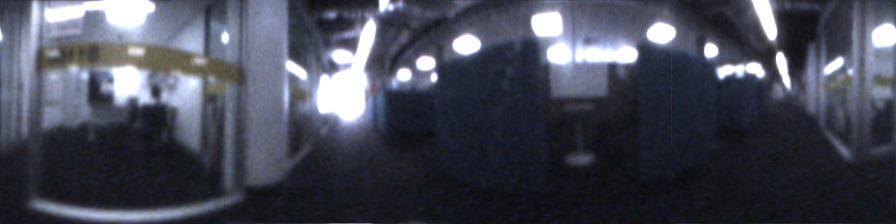}
    \end{subfigure}\hfill%
    \begin{subfigure}{0.495\linewidth}
        \includegraphics[width=1.0\linewidth]{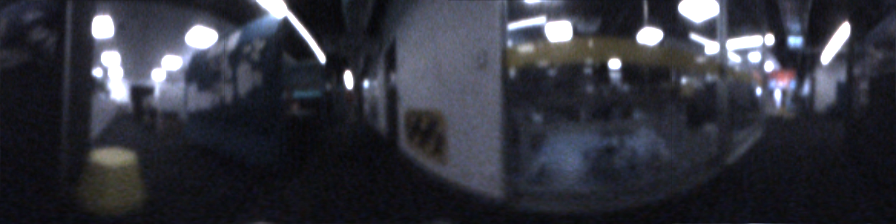}
    \end{subfigure}

    \begin{subfigure}{0.495\linewidth}
        \includegraphics[width=1.0\linewidth]{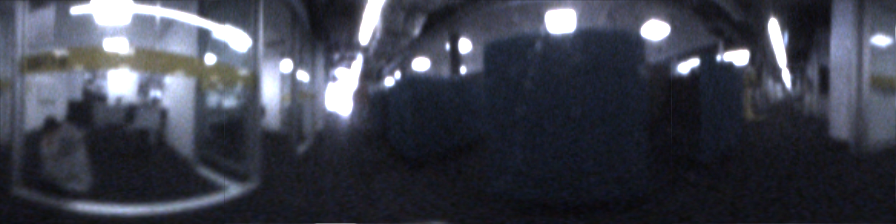}
        \caption{Different locations}
        \label{fig:aliased}
    \end{subfigure}\hfill%
    \begin{subfigure}{0.495\linewidth}
        \includegraphics[width=1.0\linewidth]{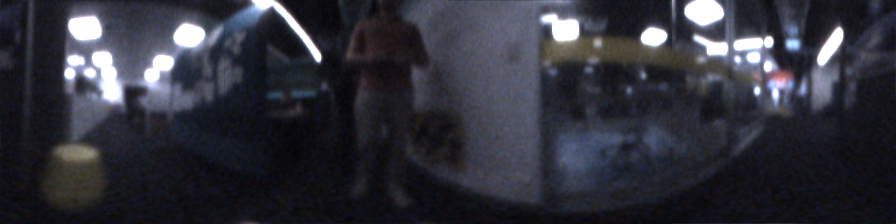}
        \caption{Same location}
        \label{fig:different}
    \end{subfigure}
    \caption{Example imagery from the recorded environment; a) two physically different locations that are visually similar, and
    b) two views of the same physical location with visual differences.}
    \label{fig:variation}
\end{figure}

Our experiments consider a mobile robot navigation task in which the robot spawns at a random location
in the environment and must reach a pre-designated goal location by executing actions from
the set \{\verb|turn_left|, \verb|turn_right|, \verb|move_forward|\}. The goal
location is fixed for all experiments. Unlike previous work on goal-driven navigation in synthetic environments~\cite{mirowski2016learning}, the goal location is not visible on the images. The agent learns to recognize the environmental landmarks solely from the photographic inputs.
The training environment was constructed from a single traversal of an office environment
on a Pioneer 3DX mobile robot with a 360-degree camera (see Fig.~\ref{fig:pioneer},
discretizing position and orientation to $\Delta_\text{pos}=10$cm and $\Delta_\text{rot}=90\degree$.
The validation traversal was collected in the same environment on a different day,
and the route was traversed in a different order. As a result, the
two datasets exhibit typical day-to-day differences in the appearance of an office
environment, including moderate variations in lighting and furniture placement, and variations
in dynamic obstacles such as humans (see Fig.~\ref{fig:variation}).

The navigation problem consists of each worker spawning at a random node and orientation in
the environment at the beginning of each episode, and choosing actions according to its policy until
$T_\text{max}=3000$ timesteps have elapsed. A node in the graph is chosen at random as the
goal (reward of $1.0$), and $N_\text{a}=10$ small sub-goals (reward of $0.1$)
are placed at random to encourage exploration, as
in~\cite{mirowski2016learning}; both types of goal locations were chosen arbitrarily without researcher input and are fixed for the entire duration of the experiment.
Reaching the goal causes the agent to re-spawn in a new random location. Agents were trained for a
total of $300$ million frames of virtual experience,
which is many orders of magnitude more experience than most robots ever obtain in the real world.

We evaluate absolute performance as measured by total reward achieved
per episode by the worst-performing worker in the pool, denoted $R_\text{min}$. The
metric $R_\text{min}$ is important for judging whether the agent has learned to \emph{reliably} navigate
the environment: a useful robot should be able to reach its goal from any location.
The pose graph recorded for these experiments consisted of
$572$ nodes, and the longest path in the graph was $227$ timesteps. Given this length
and $T_\text{max}$ of $3000$, an optimal agent should be able to achieve worst case $R_\text{min} \geq 13$.
We include as a baseline the performance of a random walk with near-optimal action prior, using
turn probability equal to the fraction of nodes in the graph representing intersections (locations where rotation
may be necessary). Note that a uniform action prior achieves negligible reward ($<0.1$), which further supports the
use of the bootstrap sampling technique.

\begin{figure}[t]
    \centering
    \begin{subfigure}{0.2\linewidth}
        \includegraphics[width=1.0\linewidth]{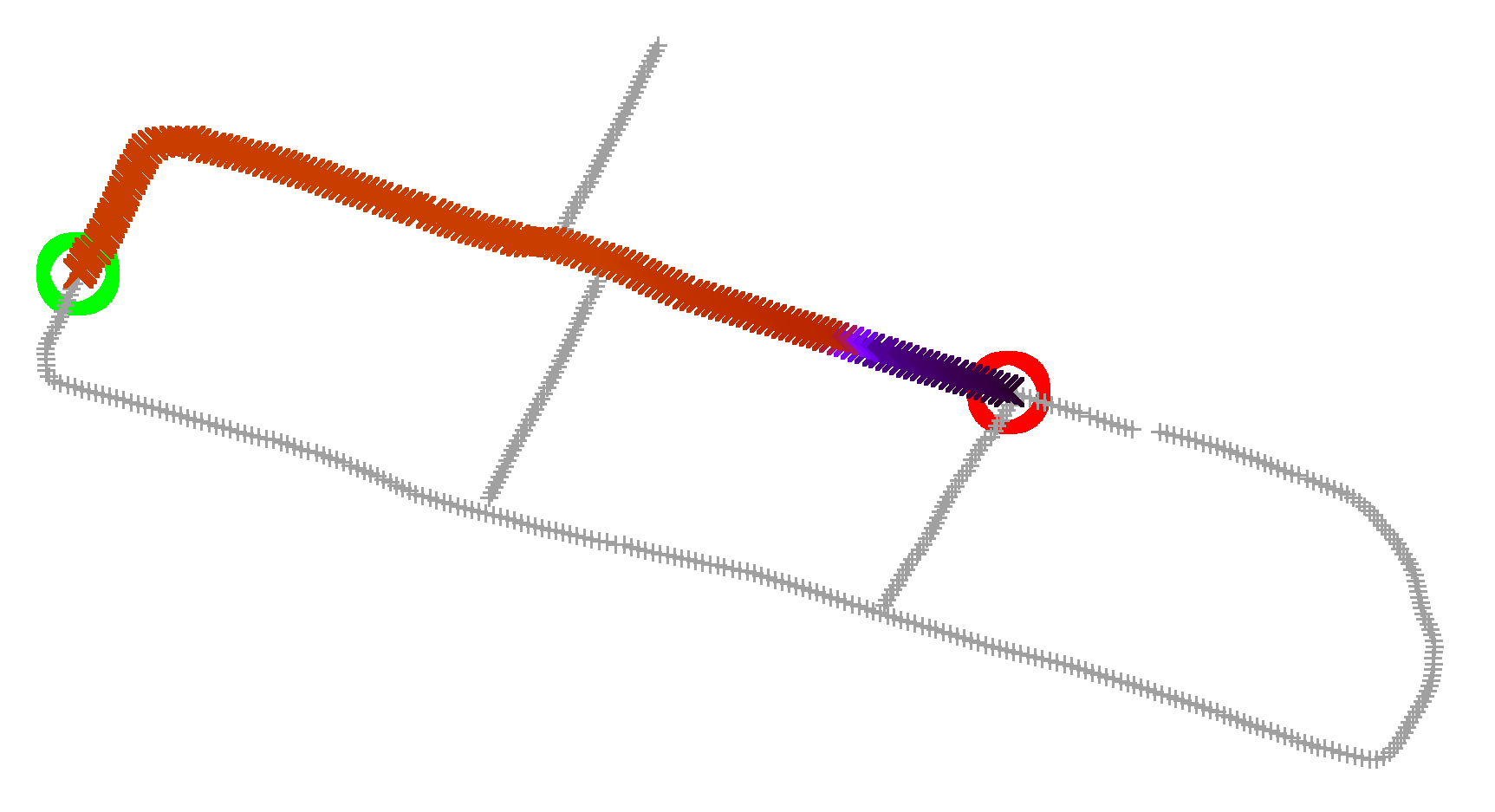}
    \end{subfigure}\hspace{4em}%
    \begin{subfigure}{0.2\linewidth}
        \includegraphics[width=1.0\linewidth]{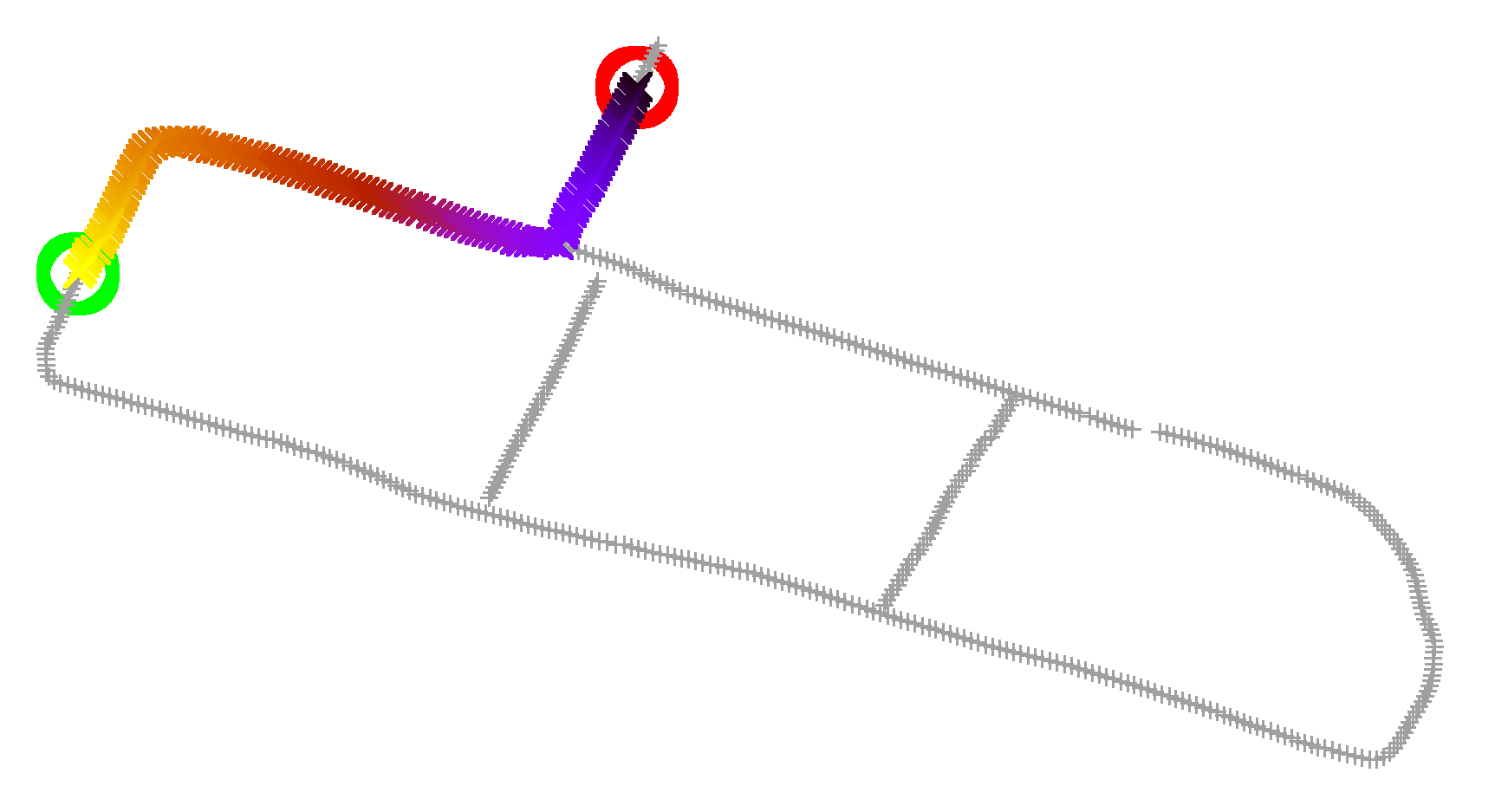}
    \end{subfigure}

    \begin{subfigure}{0.2\linewidth}
        \includegraphics[width=1.0\linewidth]{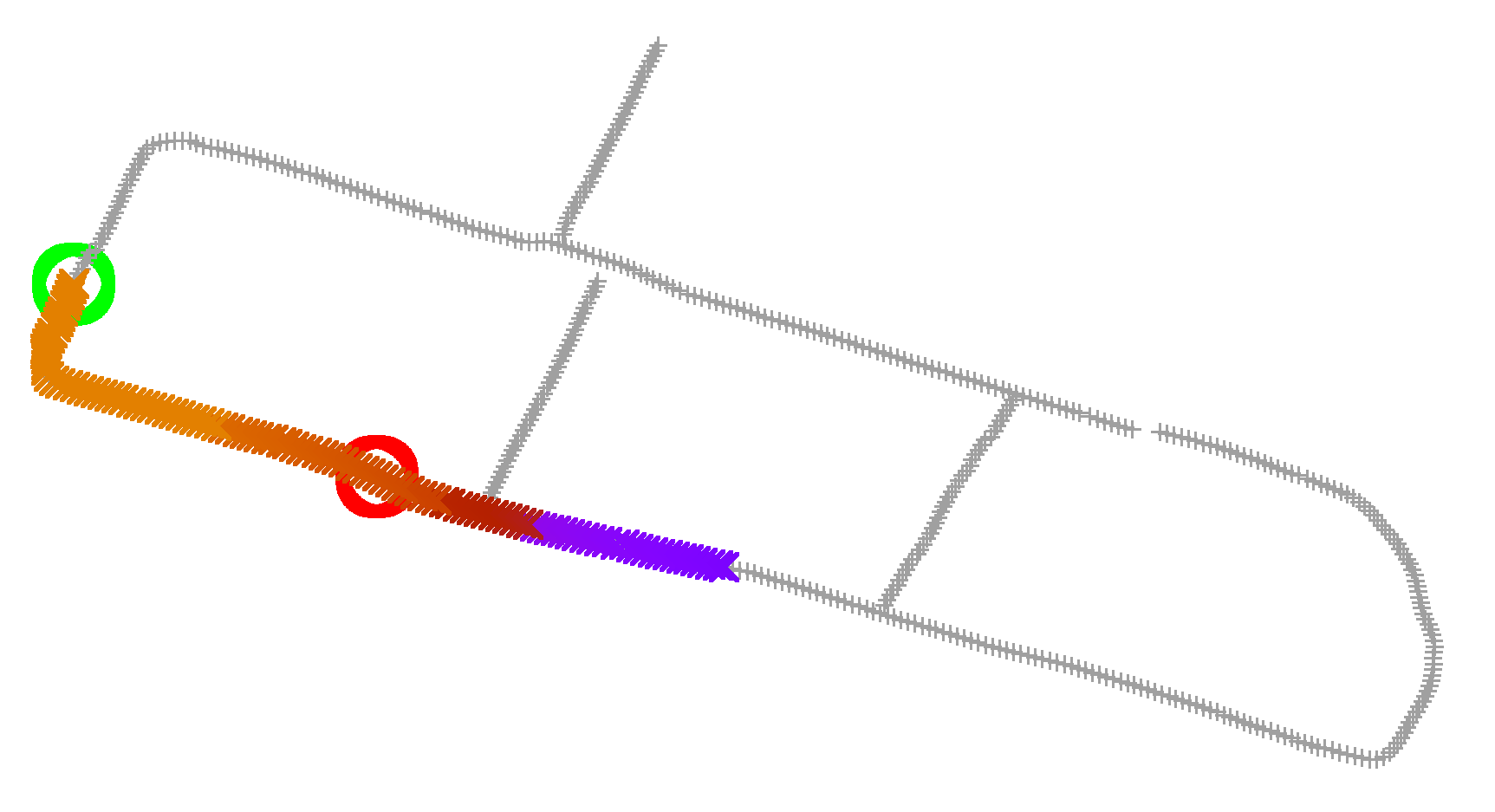}
        \caption{Sub-optimal}
        \label{fig:failure}
    \end{subfigure}\hspace{4em}%
    \begin{subfigure}{0.2\linewidth}
        \includegraphics[width=1.0\linewidth]{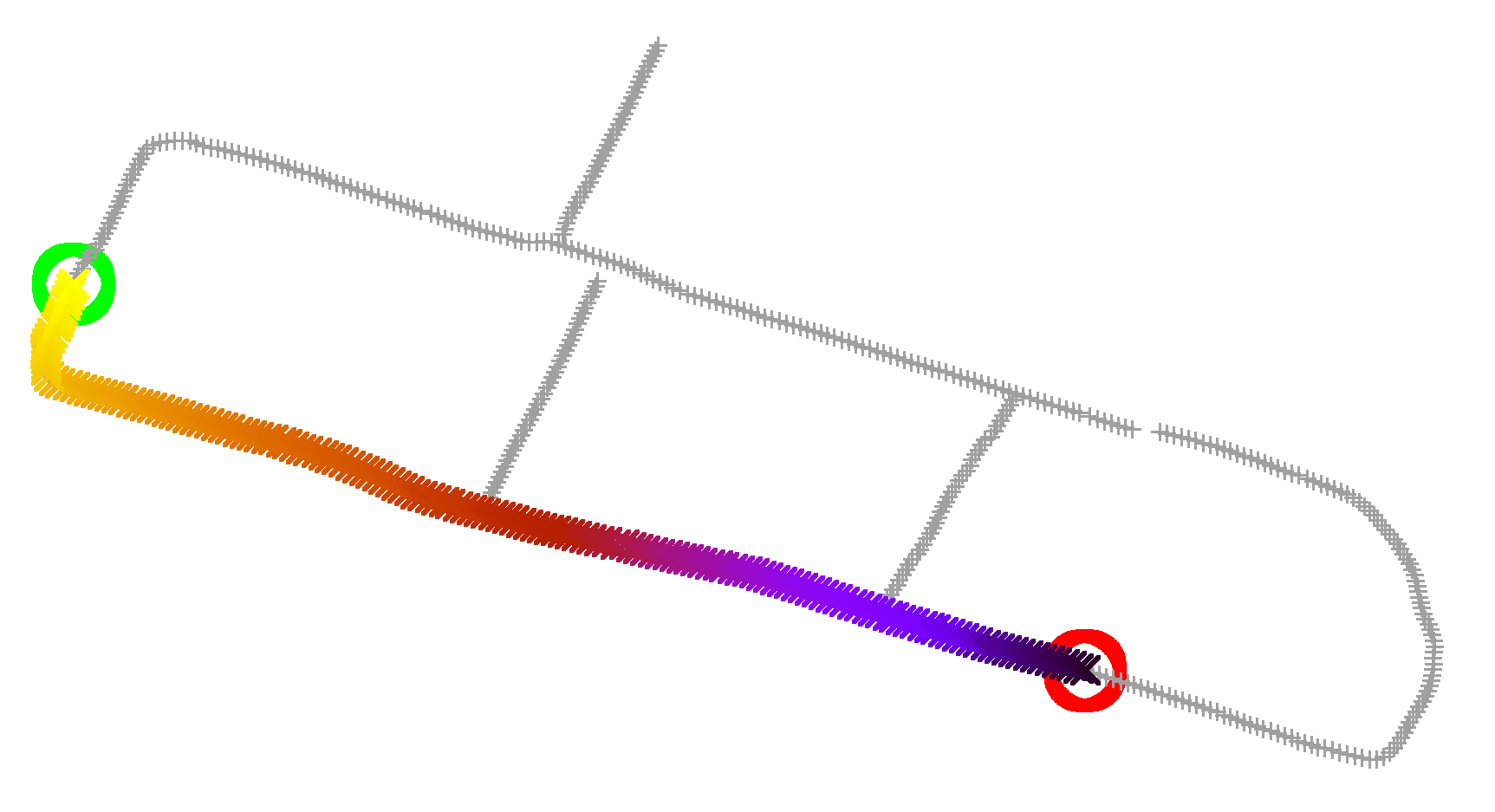}
        \caption{Directly to the goal}
        \label{fig:success}
    \end{subfigure}
    \caption{Example trajectories through the validation environment by a trained agent. Path color indicates
    progressive timesteps (purple to yellow), and red and green circles indicate start and goal locations respectively.
    a) Two sub-optimal trajectories in which the agent got stuck or backtracked but ultimately reached the goal,
    and b) two successful trajectories directly to the goal.}
    \label{fig:traj}
\end{figure}

\begin{figure}
    \centering
    \begin{subfigure}{0.475\linewidth}
        \centering
        \includegraphics[width=\linewidth]{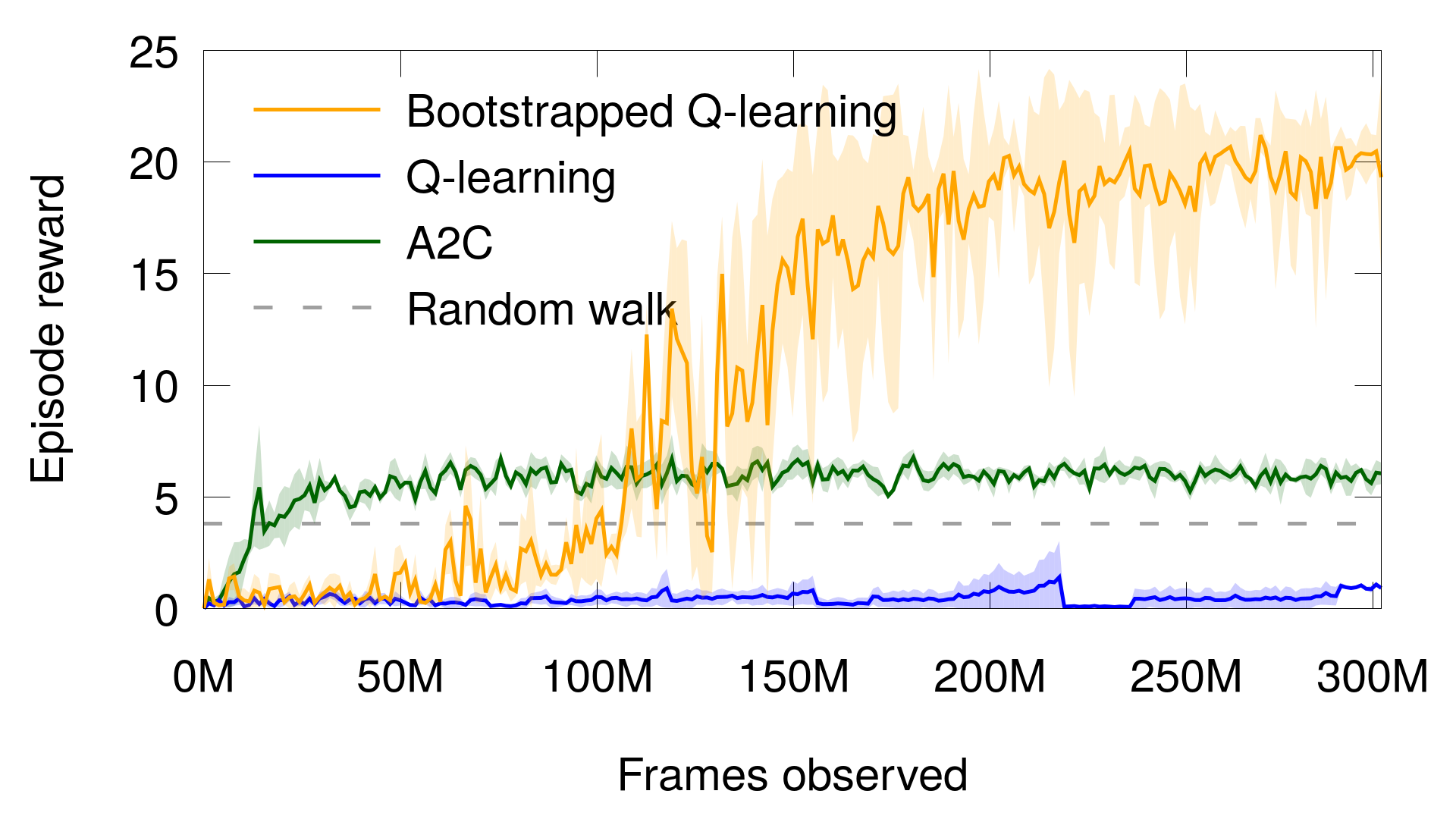}
        \caption{Mean total reward on the validation environment}
        \label{fig:alg_mean_val}
    \end{subfigure}\hfill%
    \begin{subfigure}{0.475\linewidth}
        \centering
        \includegraphics[width=\linewidth]{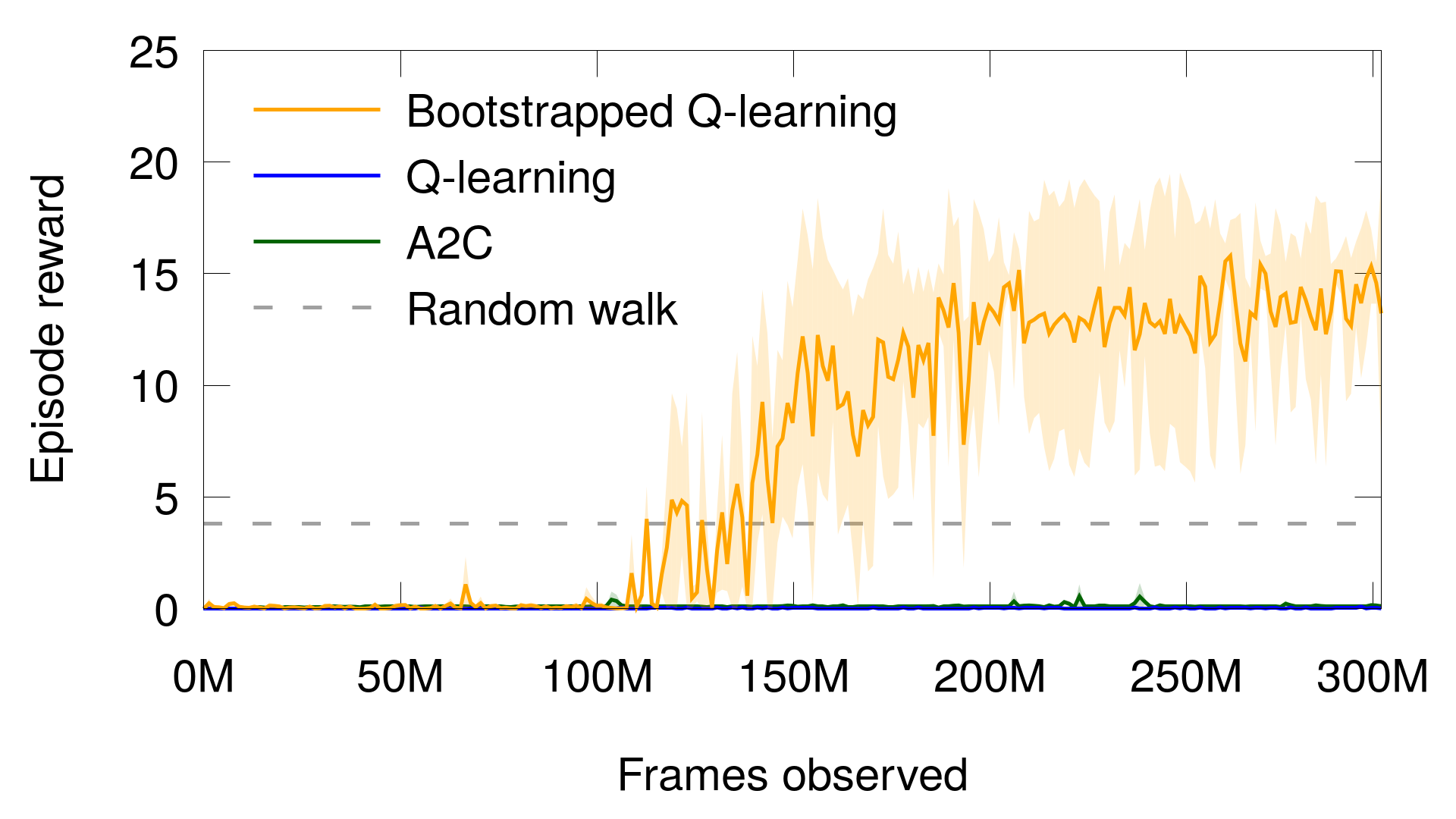}
        \caption{$R_\text{min}$ on the validation environment}
        \label{fig:alg_min_val}
    \end{subfigure}
    \begin{subfigure}{0.49\linewidth}
        \centering
        \includegraphics[width=\linewidth]{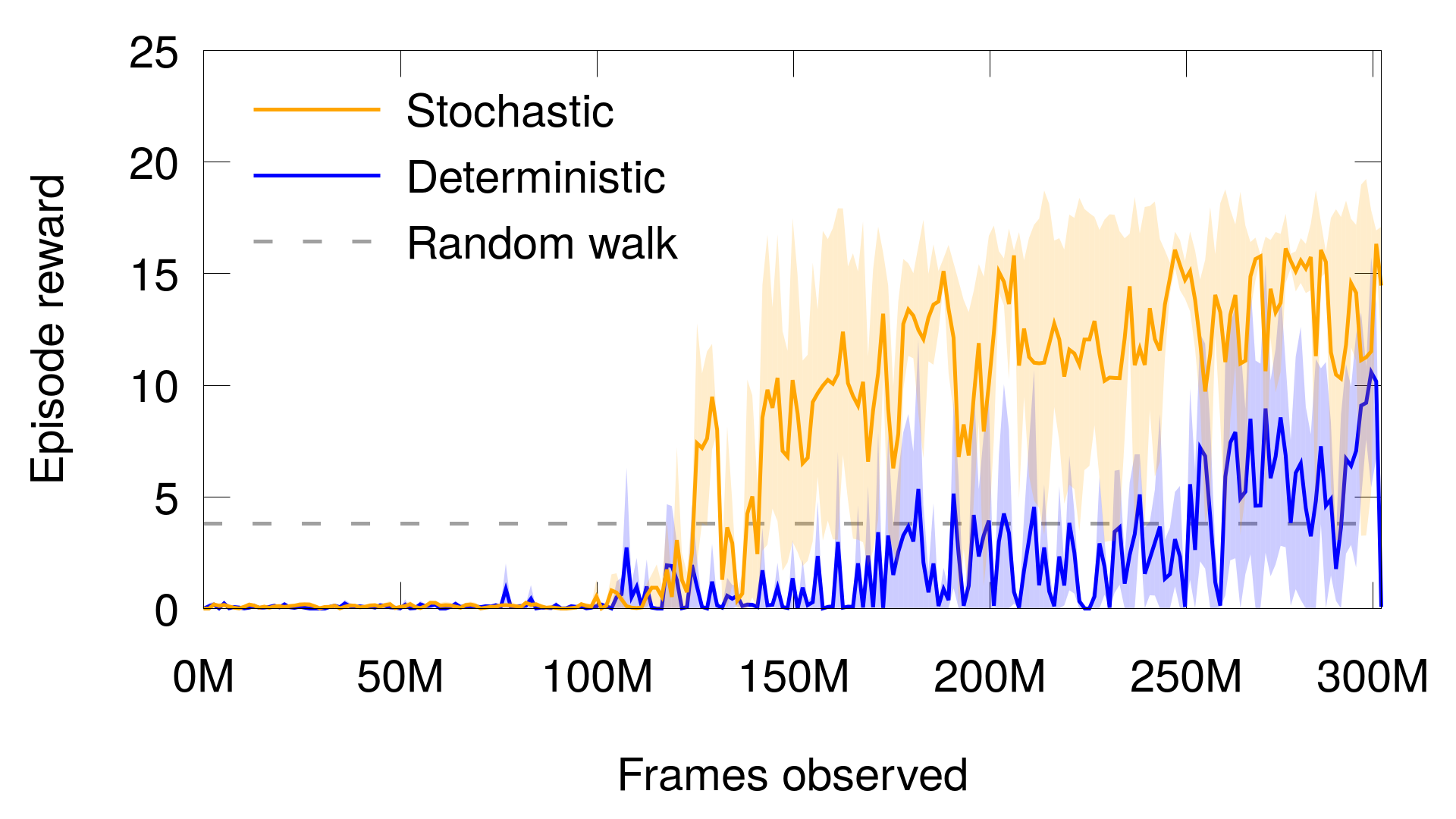}
        \caption{$R_\text{min}$ on the training environment}
        \label{fig:aug_min_trn}
    \end{subfigure}\hfill%
    \begin{subfigure}{0.49\linewidth}
        \centering
        \includegraphics[width=\linewidth]{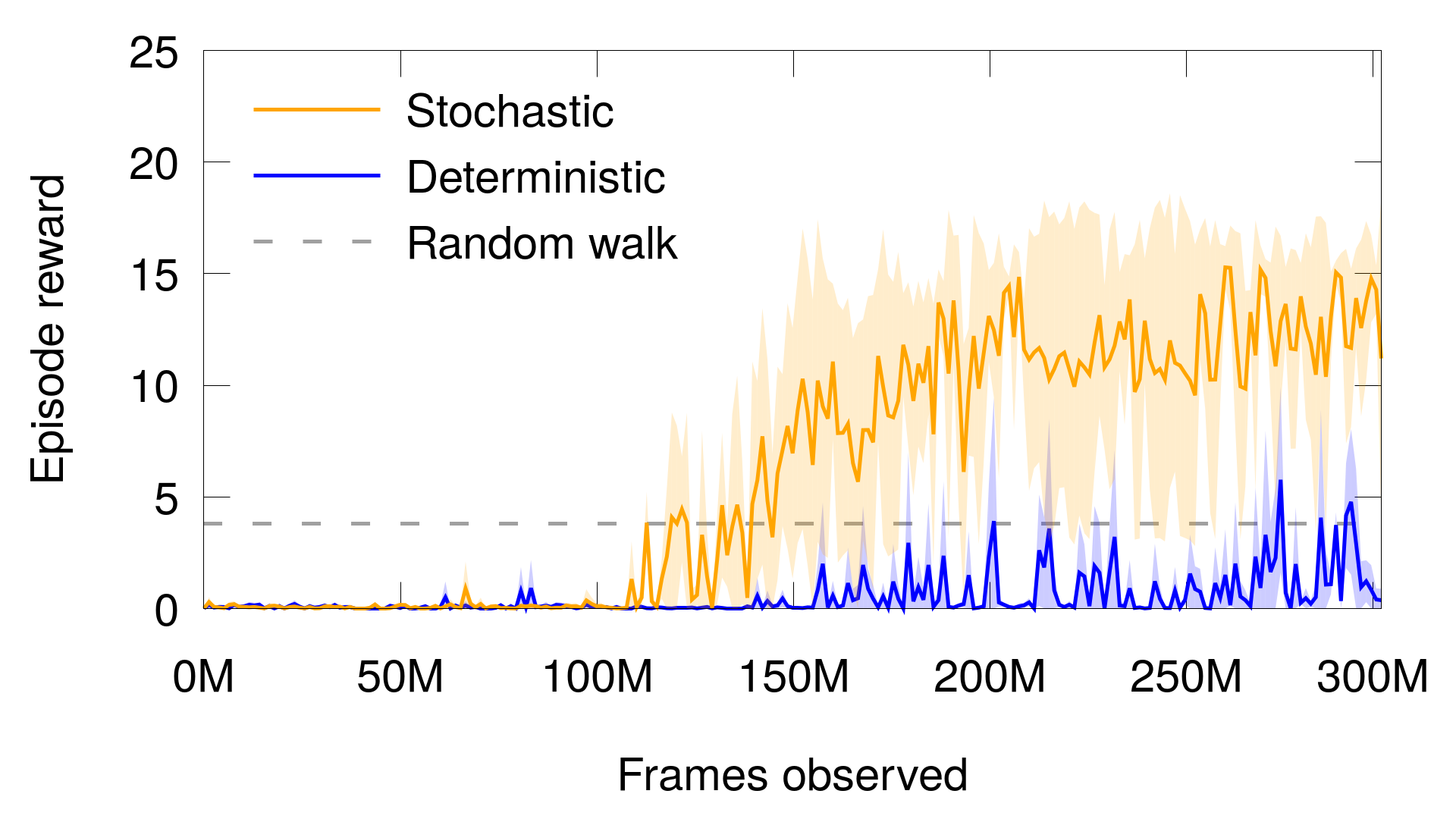}
        \caption{$R_\text{min}$ on the validation environment}
        \label{fig:aug_min_val}
    \end{subfigure}
    \begin{subfigure}{0.49\linewidth}
        \centering
        \includegraphics[width=\linewidth]{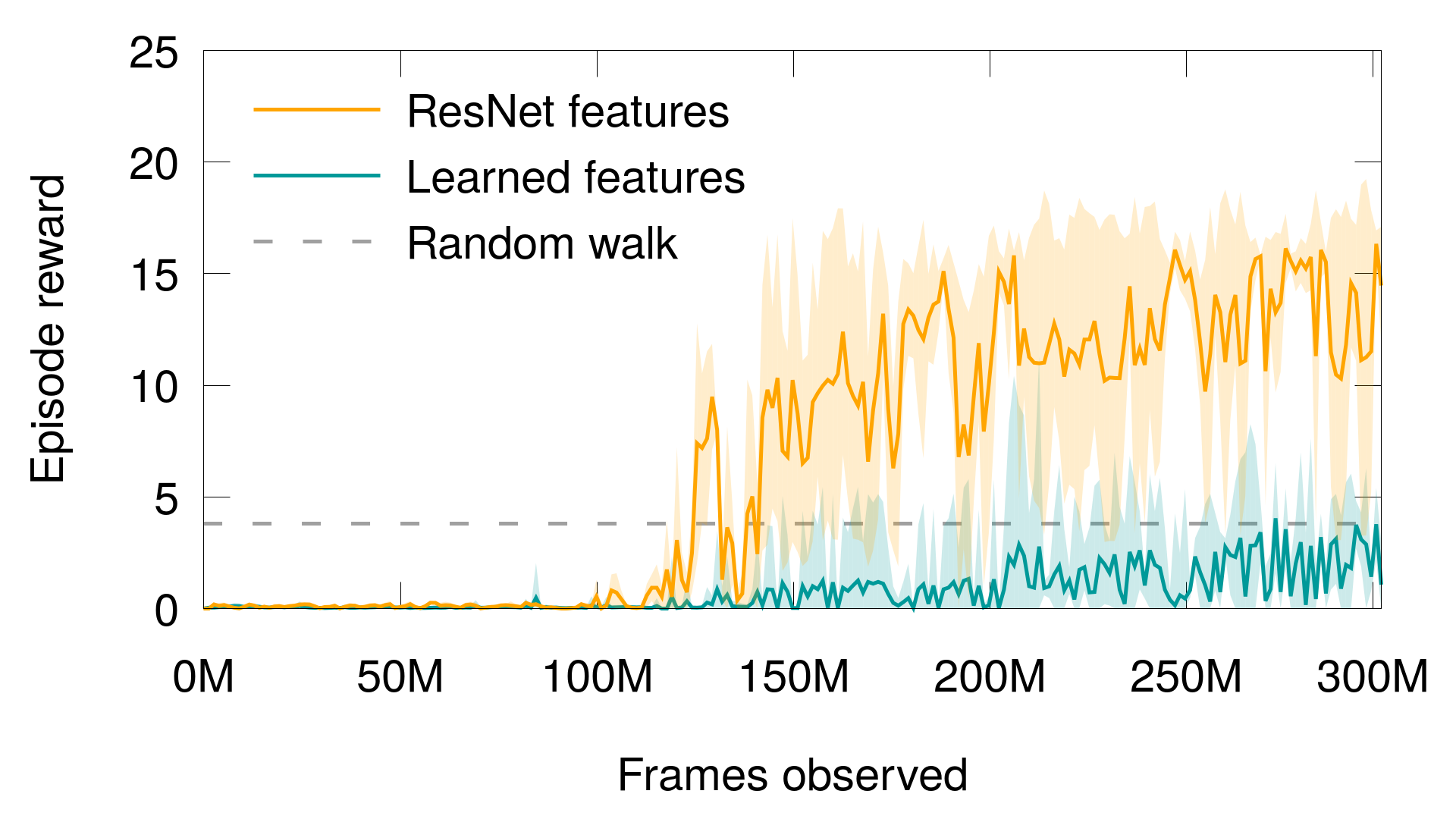}
        \caption{$R_\text{min}$ on the training environment}
        \label{fig:env_min_trn}
    \end{subfigure}\hfill%
    \begin{subfigure}{0.49\linewidth}
        \centering
        \includegraphics[width=\linewidth]{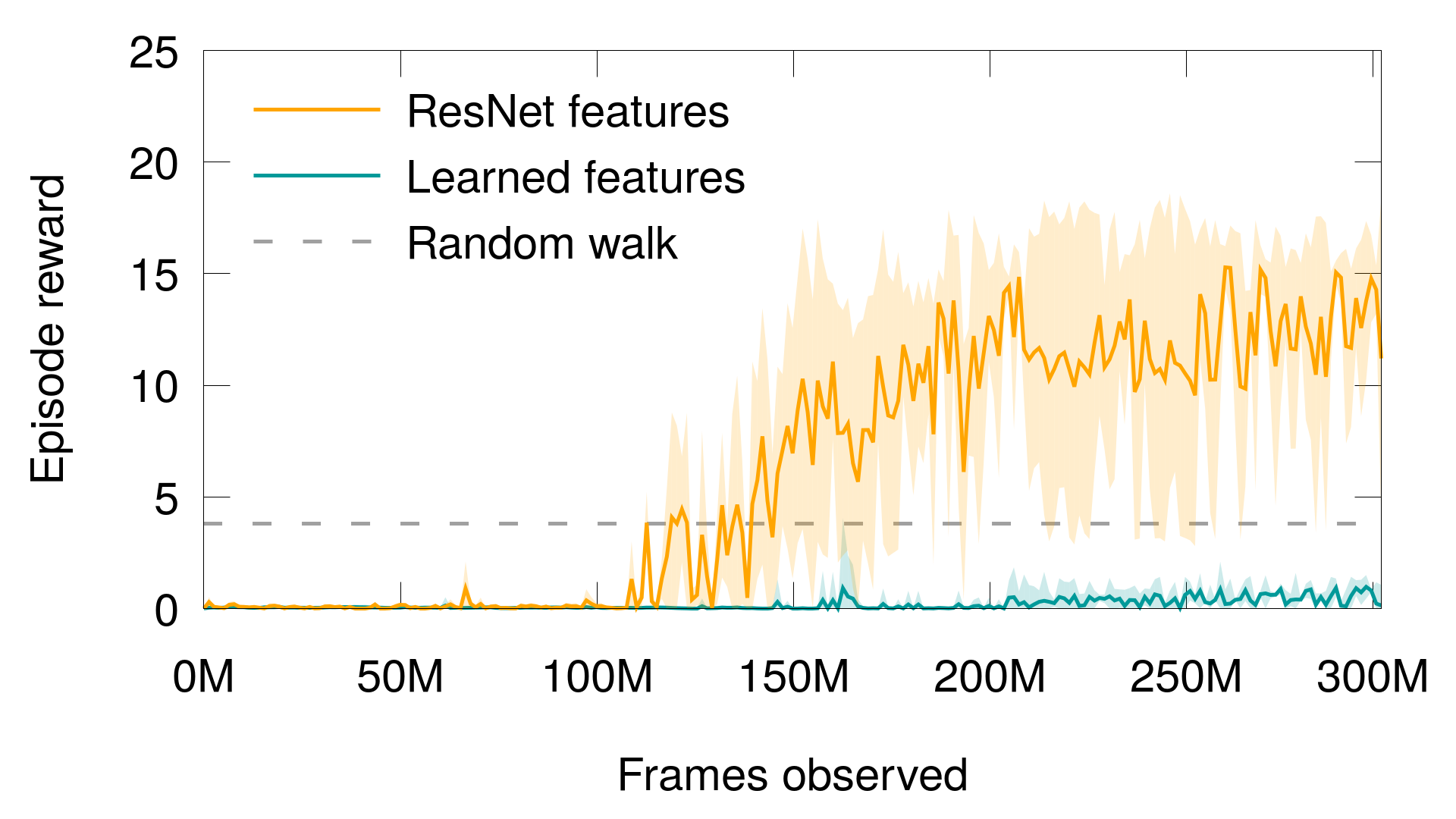}
        \caption{$R_\text{min}$ on the validation environment}
        \label{fig:env_min_val}
    \end{subfigure}
    \begin{subfigure}{0.49\linewidth}
        \centering
        \includegraphics[width=\linewidth]{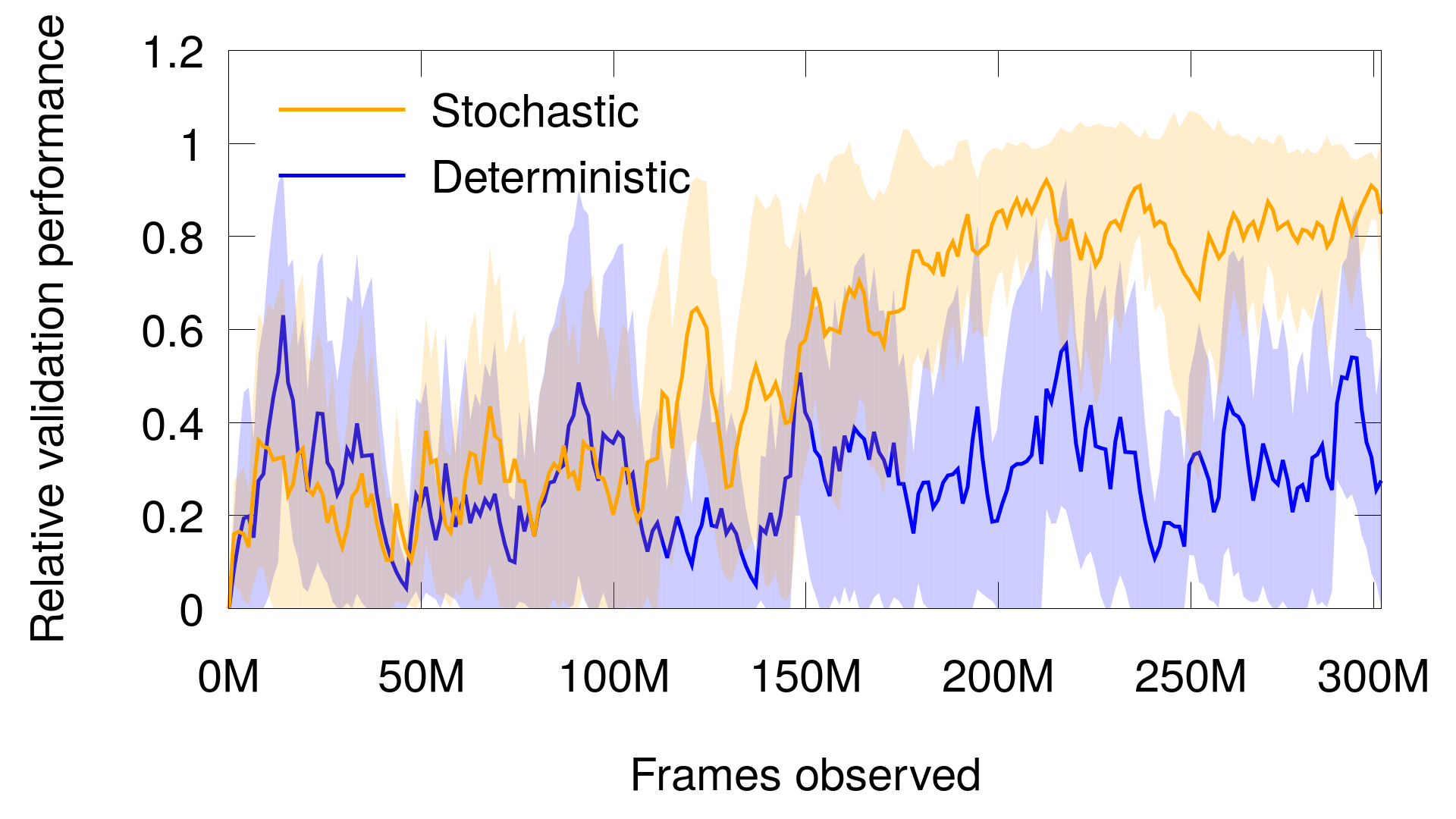}
        \caption{Transfer benefit of stochastic observations}
        \label{fig:rel_aug}
    \end{subfigure}\hfill%
    \begin{subfigure}{0.49\linewidth}
        \centering
        \includegraphics[width=\linewidth]{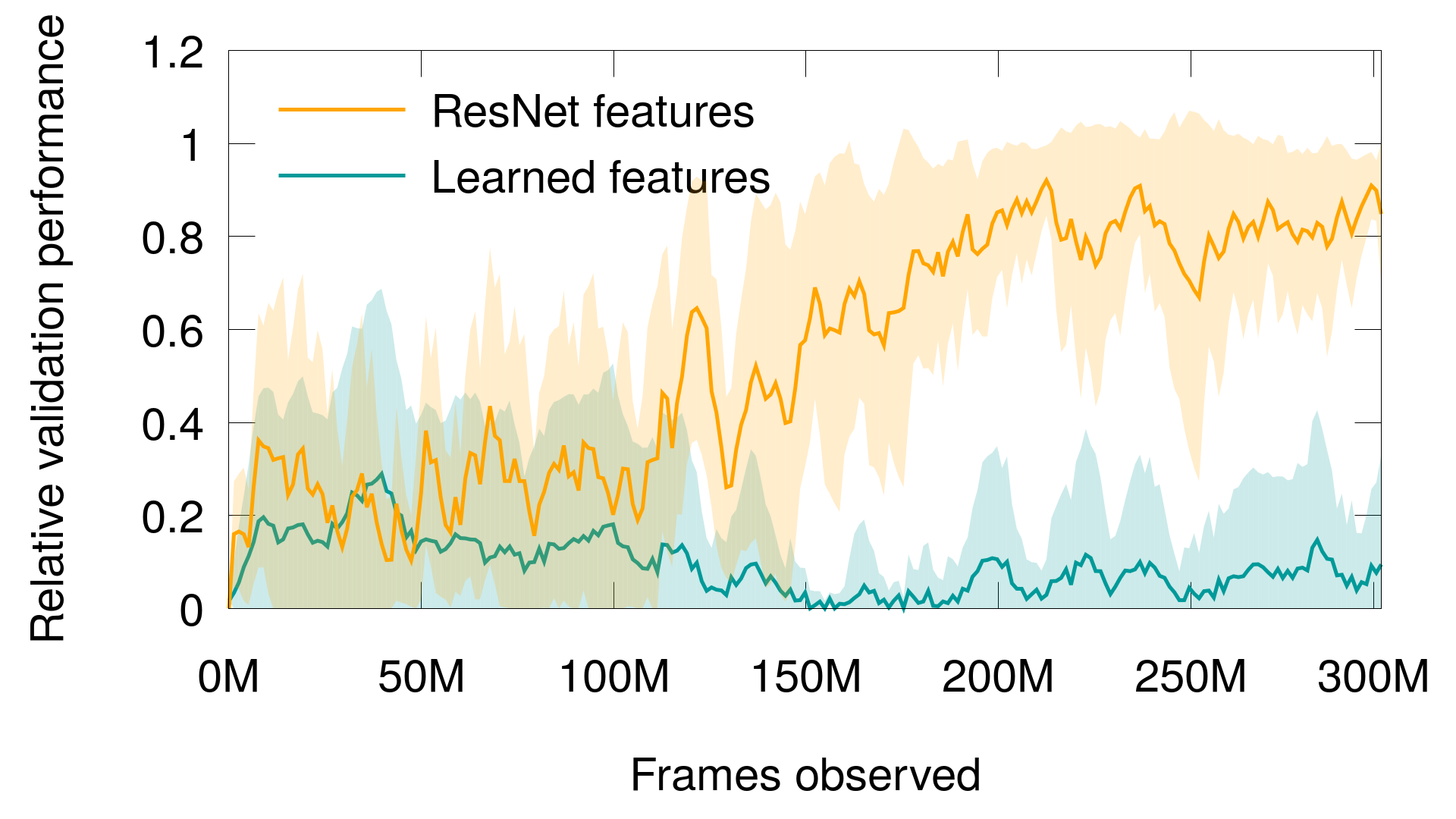}
        \caption{Transfer benefit of pre-trained visual features}
        \label{fig:rel_env}
    \end{subfigure}
    \caption{Experimental results on the navigation task. a-f) Minimum episode score $R_\text{min}$,
    indicating worst case performance,
    g-h) ratio of validation to training performance.
    a-b) Bootstrapped Q-learning performs best; the other algorithms fail to reliably solve the task.
    c-h) Augmenting the environment with stochastic observations and using a pre-trained visual encoder dramatically improve training and transfer performance.}
    \label{fig:results}
\end{figure}

In addition to measuring reward, we compare the transfer-oriented components of the system according to $R_\text{relative}$,
the ratio of $R_\text{min}$ on the validation environment to $R_\text{min}$ on the training environment: this metric
measures the fraction of training performance achieved in validation, and is important for determining whether a particular
component of the system specifically affects transfer; results are shown in Fig.~\ref{fig:results}. Qualitatively,
successful agents learn to navigate directly to the goal; see Fig.~\ref{fig:traj} for example trajectories
from a trained agent on the validation environment.

\paragraph{RL algorithm}

Our experiments compare three types of RL algorithms that learn from parallel experience:
A2C~\cite{mnih2016asynchronous}, $n$-step Q-learning~\cite{mnih2016asynchronous}, and $n$-step
bootstrapped Q-learning~\cite{osband2016deep}. A2C (a synchronous implementation of A3C in
which parallel actions are always chosen from the same model) is a policy search method that
directly optimizes a probability distribution over actions, and n-step Q-learning is a value-based
method like the bootstrapped approach but which maintains only one Q-function estimate.
Results are shown in Figs.~\ref{fig:alg_mean_val} and~\ref{fig:alg_min_val};
the bootstrapped Q-learning algorithm achieves significantly better
performance.
For the robotics context in particular, the ensemble nature of the technique provides a significant advantage
in addition to its performance, because the ensemble can be interpreted as a distribution over
Q-values to enable reasoning about model uncertainty.


\paragraph{Visual encoding}

Learning visual features from scratch on a small dataset can result in reduced representational diversity due to a lack
of variation in visual structure in the environment. We compare performance when training
the convolutional encoder of~\cite{mnih2016asynchronous,mirowski2016learning} from scratch, against the
fixed ResNet-50 architecture pre-trained on ImageNet; results are shown in Fig.~\ref{fig:env_min_val}.
Pre-trained visual features perform better on the validation environment, incur a lower penalty in transfer performance
(see Fig.~\ref{fig:rel_env}), and are much more computationally efficient because a fixed encoder allows us to
pre-encode all of the images in our environment during training.

\paragraph{Environmental stochasticity}

Dataset augmentation is a proven technique in machine learning, and we
exploit this idea by augmenting our training environment with stochastic observations;
results are shown in Figs.~\ref{fig:aug_min_trn} and~\ref{fig:aug_min_val}. Stochastic observations significantly improve performance
on both the training and validation environments. Agents trained on deterministic observations have poor
minimum performance, and cannot be said to reliably solve the task; agents with stochastic observations
however achieve nearly optimal minimum performance as computed by an optimal path planner.
In addition, transfer performance as shown in Fig.~\ref{fig:rel_aug} is significantly
improved by stochasticity, indicating that the stochastic variation we inject into the observations
improve learning in general, and also capture some degree of the natural variation encountered between traversals.

\section{Discussion and Future Work}

We have presented a reinforcement learning approach for learning to navigate on a mobile robot from a
single traversal of the environment.
The contributions of this work consist of an interactive replay buffer technique, augmenting the training environment with stochastic observations,
and the use of pre-trained visual features.
The first contribution enables the generation of extensive training experience needed by model-free systems from real world data;
the latter two contributions significantly improve performance on both the training
and validation environments, and proved crucial for obtaining reliable zero-shot transfer to variations in
the environment unseen during training.

Future work will focus on transfer to situations with agent \emph{and} robot in the loop.
Performance on the validation set provides encouraging evidence for the feasibility of closed-loop transfer, although
several issues remain to be addressed. In particular since our actions are coarse and discrete, once an action
is selected by the agent, movement to the next node in the pose graph will need to be handled by a low-level
motion controller. In our setup this will likely be accomplished by a laser-based approximate localization
system; however, contributions in visual servoing~\cite{chaumette2016visual}, teach and repeat~\cite{paton2015s},
and image alignment techniques for robot and animal homing~\cite{kodzhabashev2015route,webb2016neural} provide
evidence that a purely visual solution is also possible.

Intriguing evidence from psychology and neuroscience suggests that animals may replay known and novel sequences
of experience during sleep, in service of consolidating memory and discovering new
routes~\cite{olafsdottir2016coordinated,gundersen2015forming}. In the long term outlook for learning in robotics,
clever re-use of recorded experience is likely to be crucial for making the best of limited interaction with
the environment, and forms of virtual interactive replay may play an important role in this process.

\bibliography{main}
\bibliographystyle{unsrt}

\end{document}